\documentclass{article}
\usepackage{ijcai24}

\usepackage{times}
\usepackage{soul}
\usepackage{url}
\usepackage[utf8]{inputenc}
\usepackage[small]{caption}
\usepackage{amsthm}
\usepackage{booktabs}
\usepackage{algorithm}
\usepackage{algorithmic}
\usepackage[switch]{lineno}
\usepackage{blindtext}
\usepackage{amsmath}   
\usepackage{hyperref} 
\usepackage{bbding}
\usepackage{pifont}
\usepackage{wasysym}
\usepackage{amssymb}
\usepackage{color} 
\usepackage{booktabs}   
\usepackage{multicol}  
\usepackage{multirow}  
\usepackage{caption}
\usepackage{graphicx}
\usepackage{diagbox}    
\usepackage{multirow}   
\usepackage{bm}  
\usepackage{subfigure}

\title{Fast One-Stage Unsupervised Domain Adaptive Person Search}

\author{
Tianxiang Cui$^1$
\and
Huibing Wang$^1$\thanks{Corresponding author}\and
Jinjia Peng$^2$\and
Ruoxi Deng$^3$\and
Xianping Fu$^1$\and
Yang Wang$^{4*}$\\
\affiliations
$^1$The College of Information and Science Technology, Dalian Maritime University, China\\
$^2$The College School of Cyber Security and Computer, Hebei University, China\\
$^3$The College of Computer Science and Artificial Intelligence, Wenzhou University, China\\
$^4$Key Laboratory of Knowledge Engineering with Big Data, Hefei University of Technology, China\\
\emails
cuitianxiang1@gmail.com,
\{huibing.wang,fxp\}@dlmu.edu.cn,
pengjinjia@hbu.edu.cn,
ruoxii.deng@gmail.com,
yangwang@hfut.edu.cn
}

\begin{document}

\maketitle

\begin{abstract}
	Unsupervised person search aims to localize a particular target person from a gallery set of scene images without annotations, which is extremely challenging due to the unexpected variations of the unlabeled domains. However, most existing methods dedicate to developing multi-stage models to adapt domain variations while using clustering for iterative model training, which inevitably increases model complexity. To address this issue, we propose a \underline{F}ast \underline{O}ne-stage \underline{U}nsupervised person \underline{S}earch (FOUS) which complementary integrates domain adaptaion with label adaptaion within an end-to-end manner without iterative clustering. To minimize the domain discrepancy, FOUS introduced an Attention-based Domain Alignment Module (ADAM) which can not only align various domains for both detection and ReID tasks but also construct an attention mechanism to reduce the adverse impacts of low-quality candidates resulting from unsupervised detection. Moreover, to avoid the redundant iterative clustering mode, FOUS adopts a prototype-guided labeling method which minimizes redundant correlation computations for partial samples and assigns noisy coarse label groups efficiently. The coarse label groups will be continuously refined via label-flexible training network with an adaptive selection strategy. With the adapted domains and labels, FOUS can achieve the state-of-the-art (SOTA) performance on two benchmark datasets, CUHK-SYSU and PRW. The code is available at \href{https://github.com/whbdmu/FOUS}{https://github.com/whbdmu/FOUS}
	
\end{abstract}

\section{Introduction}
\label{sec:intro}

The objective of the pedestrian search task is to identify and locate specific individuals within panoramic images. Initially, all pedestrians present in the image are detected, and subsequently, they are compared to the target pedestrian image for recognition. In the context of the intelligent security field, it becomes increasingly difficult and costly to complete person search tasks in a supervised manner due to the massive volume of videos and pictures generated by numerous cameras. Therefore, unsupervised domain adaptive methods should receive more attention.

\begin{figure}[tbp!]
	\centering
	\setlength{\belowcaptionskip}{-1mm}
	\vspace{-0.35cm} 
	\subfigtopskip=-1pt 
	\subfigbottomskip=-1pt 
	\subfigcapskip=-5pt 
	\subfigure[Clustering-based person search]{\label{f1a}
		\includegraphics[width=3.5in]{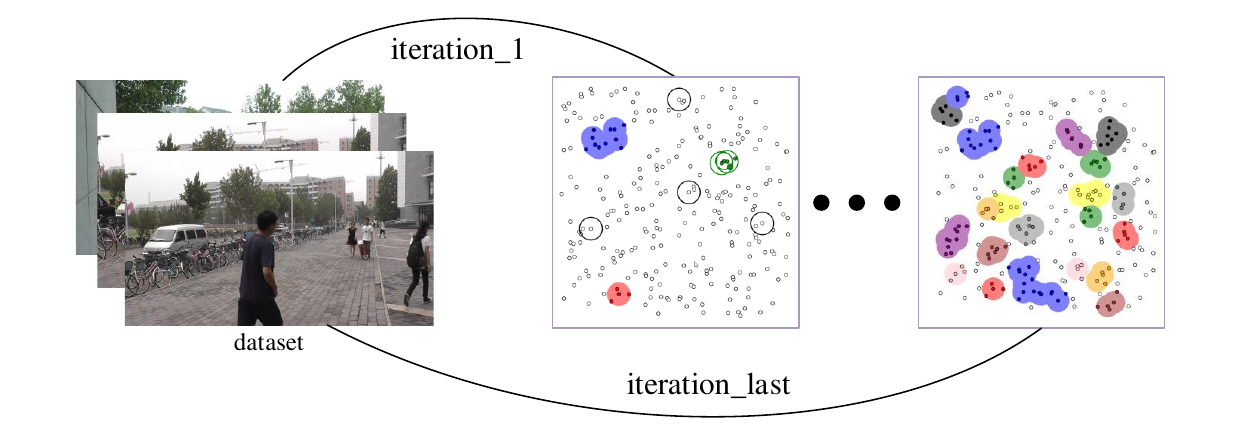}}
	\quad    
	\subfigure[Ours]{\label{f1b}
		\includegraphics[width=3.5in]{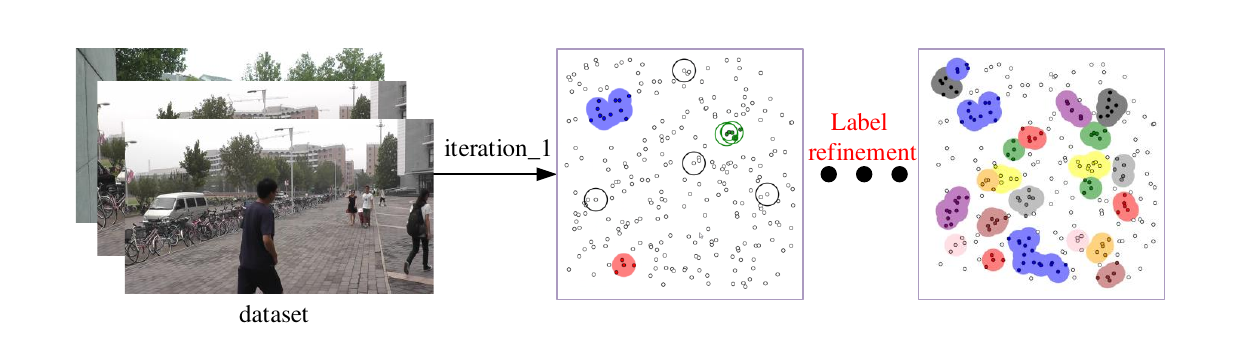}}	
	
	\caption{Differences in operation between our proposed method and mainstream methods. \subref{f1a} represents that the clustering algorithm utilizes the original data to calculate similarity in each iteration. \subref{f1b} illustrates that our proposed method utilizes the original data only in the first iteration to assign soft labels, and then gradually refines the labels. }
	\label{f1}
\end{figure}

The main challenges currently in unsupervised pedestrian search are the lack of bounding box annotations and the lack of pedestrian identity label annotations. Since it is very difficult to directly train unsupervised detection and re-identification, a series of weakly supervised research works\cite{yan2022exploring,han2021weakly,jia2023collaborative} have been proposed to address the challenge of lacking identity annotations, which have employed clustering techniques to assign pseudo labels. As shown in Fig.\ref{f1a}, clustering approaches typically require multiple iterations of calculating sample similarity, which can be time-consuming and resource-intensive. Moreover, these methods still demand significant time and effort to annotate ground-truth bounding boxes, and their model generalization capability is limited, rendering them unsuitable for real-world scenarios. 

\begin{figure*}
	\centering
	\includegraphics[width=1.0\linewidth]{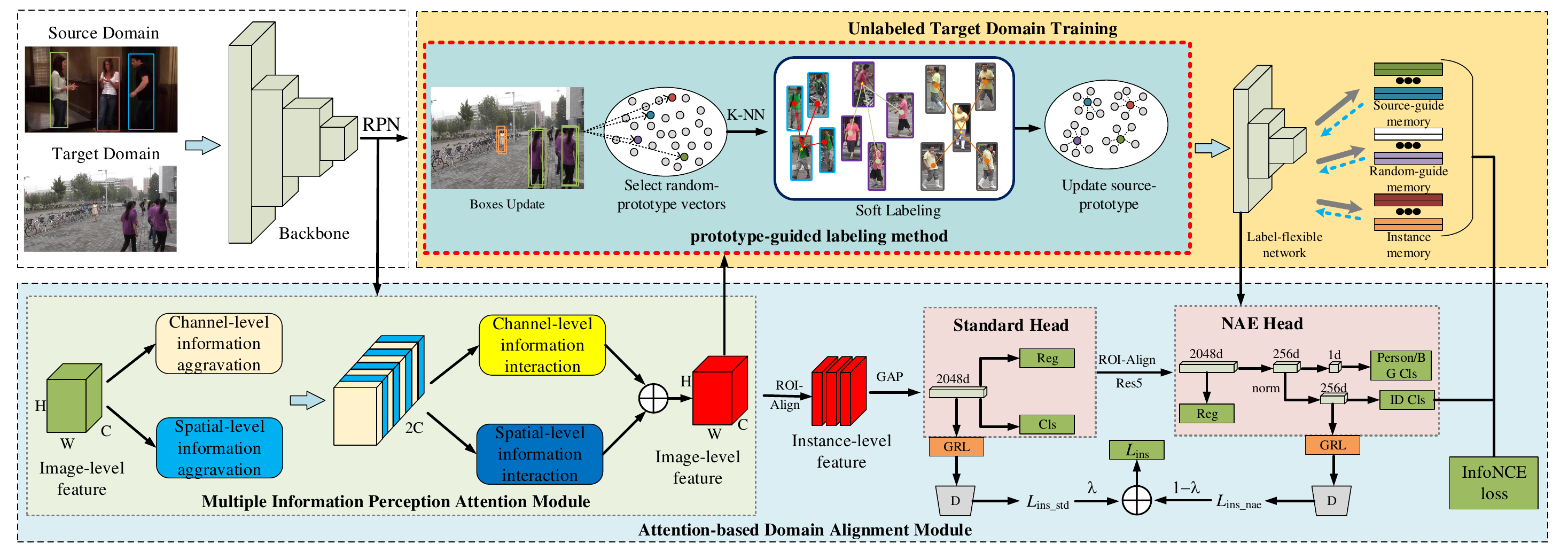}
	\caption{The design architecture of the FOUS framework. For each iteration, after extracting features, FOUS alternates between two phases: (1) Attention-based Domain Alignment.(Sec. 2.2) The candidate box quality is enhanced through the Multi-Information Perception Attention Module, followed by domain alignment operations. (2) Unlabeled Target Domain Training.(Sec. 2.3)  Annotate the unlabeled samples in the target domain utilizing prototypes as the reference for further fine-tuning, in
    which select random features as the target prototype vectors and update the learned source prototypes.}
	\label{fig:enter-label}
\end{figure*}
Therefore, unsupervised person search was emerged as the times require. Considering the difficulty and impracticality of directly training unsupervised detection and re-identification, current research methods rely on powerful pre-trained models. Li et al.[4] introduced a domain adaptive method to address the challenge of unsupervised pedestrian search by training on a labeled dataset from the source domain. Specifically, Li et al. designed a domain alignment module to alleviate the difference between unsupervised detection and re-identification subtasks, and then proposed a dynamic clustering strategy to generate pseudo labels on the target domain. We can observe that the clustering method, which is widely employed in both weakly supervised and unsupervised person re-identification tasks, offers certain advantages. It is capable of withstanding noisy labels and generating dependable labels, thus effectively training a model with robust generalization capabilities. However, the computational resources required for this method are quite demanding. The clustering method involves calculating pairwise similarities and multiple iterations to optimize results, which can result in an impractical computational burden. Consequently, these methods face challenges when applied to smart security scenarios involving a massive amount of video footage. 

In order to solve this problem, this paper proposes a fast and efficient unsupervised domain adaptation framework for the person search task, which can learn a model with strong generalization ability at a lower cost. As illustrated in Fig.\ref{f1b}, different from the clustering methods of existing models, FOUS gives up generating relatively accurate labels through complex clustering algorithms, and instead tries to learn potential correlations from soft labels with greater noise and fewer resources. First, in order to quickly assign soft labels, we use the prototype correlation labeling method to calculate the similarity between partial samples only once. To reduce computational resources, these prototypes are used as annotators to soft-label unlabeled samples based on nearest neighbor correlations. Secondly, considering the importance of mitigating the impact of generated noisy labels, we follow the domain alignment module in DAPS and introduce a new attention mechanism, which can amplify global-dimensional interaction features while reducing information dispersion. Effectively reduce the number of noisy labels generated by low-quality candidate frames in unsupervised detection. And FOUS also proposes an adaptive selection strategy to perform label-flexible training of the network to gradually refine the rough labels.

Our contributions can be summarized as follows:

1. We introduce a fast unsupervised domain-adaptive pedestrian search framework based on a prototype correlation labeling method instead of the prevalent clustering algorithm. Compared with multiple iterations of the clustering algorithm, FOUS only calculates the similarity between some samples once and directly assigns soft label groups.

2. In order to reduce the noise in the soft label group, we first introduce a new attention mechanism to alleviate the differences between unsupervised detection and re-identification tasks, and also propose an adaptive selection strategy to gradually refine the coarse labels to find reliable and similar target images.

3. Without any auxiliary labels in the target domain, our method achieves good performance on both pedestrian search benchmark datasets, which not only reduces the amount of calculation and improves the inference speed but also has good generalization ability.

\section{Method}
\label{sec:met}
\subsection{Framework Overview}

Fig.\ref{fig:enter-label} highlights the design architecture of the proposed FOUS framework. Given the input images from both the source and the target domain, we utilize a pre-trained model on the source domain annotation dataset to produce the source-prototype vector, and the image-level feature maps are extracted with a backbone network. Next, these features are fed into the Region Proposal Network (RPN) to generate proposals, which are passed through the attention module to create instance-level feature maps. To bridge the domain gaps and ensure effective detection and ReID (Re-Identification) tasks, we introduce an attention-based domain alignment module (ADAM). The ADAM is designed to align both image-level and instance-level features across different domains. It helps harmonize the representations and close the gap between domains.

To tackle the issue of computational cost in unsupervised reID tasks, a new prototype-guided labeling method is proposed to assign and optimize soft labels instead of computationally complex clustering methods, which frees up the process of label generation, including prototypes-guided labeling phase and label-flexible training phase. We first annotate the unlabeled samples with coarse labels in a simple way, which selects the prototype vectors generated above and softly annotates unlabeled samples via the relevance between these prototypes and samples. Additionally, FOUS trains its models using a combination of cluster-level contrastive loss and instance-level loss, which ensures that the models adhere to both the constraints imposed by the two types of losses. It aims to refine the coarse labels by incorporating the proposed adaptive selection strategy. Above all, except for the initial iteration,  the labeled samples from the source domain are not utilized in our FOUS.

\begin{figure}[tbp!]
	\centering
	\includegraphics[width=1.0\linewidth]{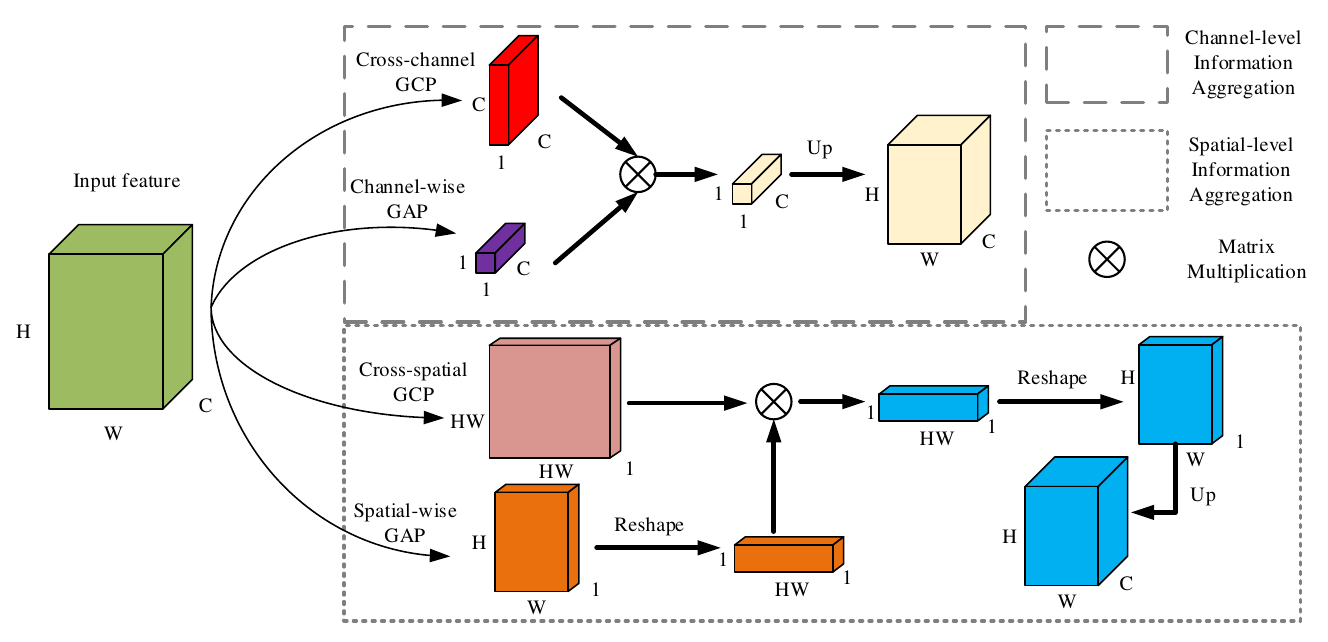}
	\caption{The channel-level information aggregation and spatial-level information aggregation in ADAM.}
	\label{f3}
\end{figure}

\subsection{Attention-based Domain Alignment Module}

\textbf{Attention Mechanism.} To mitigate the negative effects of low-quality proposals that arise from unsupervised detection, a novel attention module is designed, which can amplify global-dimensional interactive features while reducing information dispersion. Fig.\ref{fig:enter-label} emphasizes that the attention module primarily consists of information aggregation and information interaction.

Input feature $x\in \mathbb{R} ^{H\times W\times C} $ is initially processed into feature $x^c\in \mathbb{R} ^{H\times W\times C} $ and feature $x^s\in \mathbb{R} ^{H\times W\times C} $ through channel-level information aggregation and spatial-level information aggregation respectively. And these two features are subsequently cross-connected in the channel dimension to yield features $x^{cs}\in \mathbb{R} ^{H\times W\times 2C} $. The information aggregation is formulated as:
\begin{equation}
	x^{cs} = Concat(f_{clia}(x), f_{slia}(x)),
\end{equation}
where $f_{clia}$, $f_{slia}$ and $Concat$ represent channel-level information aggregation, spatial-level information aggregation and cross-concatenation, respectively. Channel-level information aggregation can capture channel dependency information, channel structure information, and spatial global information through cross-channel global covariance pooling and channel-level global average pooling. Spatial-level information aggregation can capture spatial dependency information, spatial structure information, and channel global information through cross-spatial global covariance pooling and spatial-level global average pooling. Cross-concatenation plays an important role in organizing various pieces of information, enabling effective interaction and integration of subsequent information.

Fig.\ref{f3} illustrates the specific structure of channel-level information aggregation and spatial-level information aggregation in ADAM, which can simultaneously achieve the perception of multi-dimensional dependency information, multi-dimensional global information and multi-dimensional structural information in information aggregation. The channel-level information aggregation is formulated as follows:
\begin{equation}
    {f_{clia}}(x) = Up\left( {MM\left( {CcGCP(x),CwGAP(x)} \right)} \right)
\end{equation}
where $CcGCP$, $CwGAP$, $MM$ and $Up$ represent cross-channel global covariance pooling, channel-wise global average pooling, matrix multiplication and upsampling operations, respectively. Moreover, the spatial-level information aggregation is formulated as follows:
\begin{equation}
    {f_{slia}}(x) = Up\left( {MM\left( {SwGAP{{(x)}^\Delta },CsGCP{{(x)}^\Delta }} \right)} \right)
\end{equation}
where $SwGAP$, $CsGCP$ and $\Delta$ represent spatial-wise global average pooling, cross-spatial global covariance pooling and reshape operations, respectively.

Similarly, feature $x^{cs}\in \mathbb{R} ^{H\times W\times 2C} $ is separated into feature $x^{c^{\prime }s}\in \mathbb{R} ^{H\times W\times C} $ and feature $x^{cs^{\prime }}\in \mathbb{R} ^{H\times W\times C} $ through channel-driven information interaction and spatial-driven information interaction respectively. And these two features are subsequently adaptively fused into features $x^{c^{\prime }s^{\prime }}\in \mathbb{R} ^{H\times W\times C} $ by allocating learnable parameters $\alpha\in \mathbb{R} ^{1\times 1\times C} \mathbb{} $ and $\beta\in \mathbb{R} ^{1\times 1\times C} $. The information interaction is formulated as:
\begin{equation}
	x^{c^{\prime }s^{\prime }} = \alpha f_{cdii}(x^{cs}) + \beta f_{sdii}(x^{cs}),
\end{equation}
where $f_{cdii}$ and $f_{sdii}$ represent channel-driven information interaction and space-driven information interaction, respectively. Channel-driven interaction focuses on perceiving diversity in channels by assigning unique parameters along the channel dimension while sharing the same parameters in the spatial dimension. On the other hand, spatial-driven interaction aims to perceive diversity in space by assigning distinct parameters along the spatial dimension while sharing the same parameters in the channel dimension.

\begin{figure}[tbp!]
	\centering
	\includegraphics[width=1.0\linewidth]{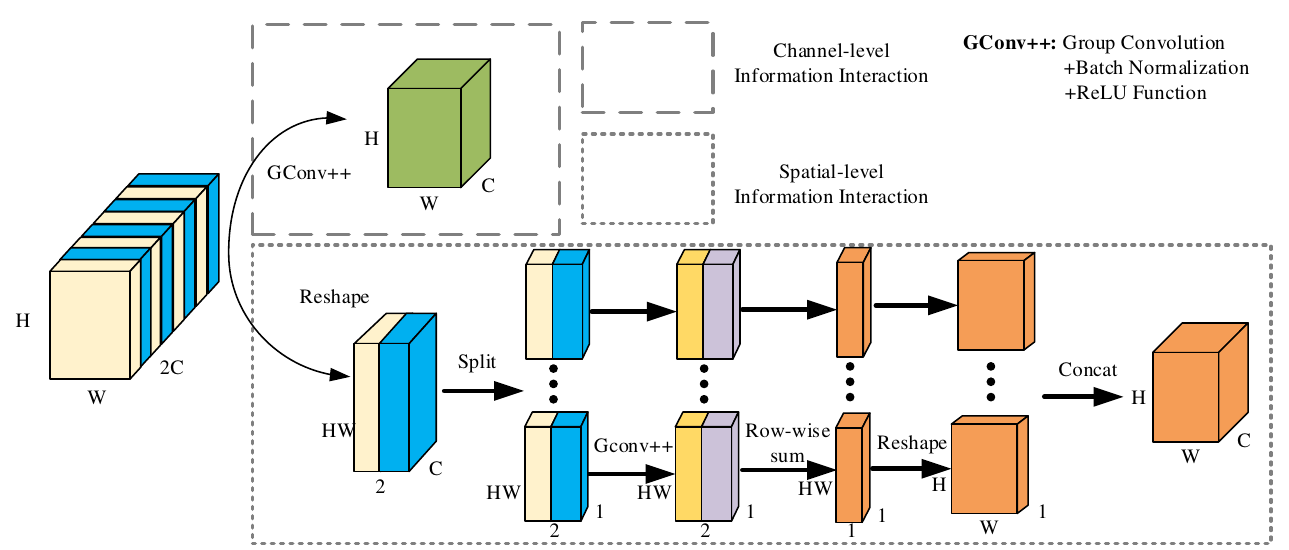}
	\caption{The channel-level information interaction and spatial-level information interaction in ADAM.}
	\label{f3_2}
\end{figure}

Fig.\ref{f3_2} shows the specific structure of channel-level information interaction and spatial-level information interaction in ADAM, which concurrently achieves the perception of multidimensional diverse information during information interaction. ADAM stimulates intrinsic information potential through a more comprehensive proactive interaction. The channel-level information interaction is formulated as follows:
\begin{equation}
	{f_{cdii}}({x^{cs}}) = GCon{v_{ +  + }}({x^{cs}})
\end{equation}
where $GConv$ represents the combination of group convolution, batch normalization, and the ReLU function. The information interaction between different groups is undisturbed. Channel-driven information interaction, achieved by assigning different parameters in the channel dimension and sharing the same information parameters in the spatial dimension, not only facilitates the interaction and fusion of multiple information in channel-level and spatial-level information collection but also further perceives channel diversity information. 

The spatial-level information interaction is formulated as follows:
\begin{equation}
	{f_{sdii}}({x^{cs}}) = Concat\left( {Sum{{\left( {GCon{v_{ +  + }}(Split({x^{cs\Delta }}))} \right)}^\Delta }} \right)
\end{equation}
where $Sum$ represents the row-wise summation. Reshaping and splitting operations are used to adjust the feature shape for subsequent specific information interactions. The spatial-driven information interaction, achieved by assigning different parameters in the spatial dimension and sharing the same parameters in the channel dimension, not only facilitates the interaction and fusion of multiple information from channel-level and spatial-level information collection but also further perceives spatial diversity information.

Finally, feature $x^{c^{\prime }s^{\prime }}\in \mathbb{R} ^{H\times W\times C}$ is activated into the attention map by the sigmoid function. The attention map is employed to modulate input sub-feature $x\in \mathbb{R} ^{H\times W\times C}$, resulting in the output sub-feature $x^{\prime}\in \mathbb{R} ^{H\times W\times C}$:
\begin{equation}
	x^{\prime} = x \mathcal{S}{igmoid}(x^{c^{\prime }s^{\prime }})
\end{equation}
The sigmoid function distinguishes the importance by activating eigenvalues within the range of 0 to 1. Importantly, the input sub-features are transformed into output sub-features across all branches. This multi-branch structure is advantageous in activating different attention mechanisms, allowing for the selective perception of valuable feature information on each branch in a targeted manner.

\textbf{Task-sensitive Domain Alignment.}
As studied by \cite{chen2018domain,you2019universal,xu2020exploring,wang2024unpacking}, reducing domain disparities is advantageous for both subtasks within person search, and can enhance the model's capacity to learn domain-invariant representations. Given the swift advancement of domain-adaptive detectors\cite{khodabandeh2019robust,wang2021survey,vibashan2023instance,zhang2023detr}, FOUS proposes a task-sensitive domain alignment module. As shown in Fig.\ref{fig:enter-label}, ADAM uses a patch-based domain classifier to predict the domain from which the input features come, and uses a minmax formula to mislead the domain classifier and encourage domain-invariant representation learning. Assuming we possess $N$ training images $\left \{I_1,...,I_N \right \} $ along with their respective domain labels, where $0$ and $1$ represent the source domain and target domain respectively. Then denote the domain classifier as $D$, and further denote the domain prediction result of the input feature $I_i$ as $p_i$. We incorporate a cross-entropy loss for domain alignment through an adversarial training approach:
\begin{equation}
	L_{dom} = -\sum_i [d_i logp_i + (1-d_i) log(1-p_i)]
\end{equation}

Based on the characteristics of the upstream and downstream tasks, we suggest achieving balance in the task-sensitive alignment module by equitably adjusting the alignment weights of instance-level features in both sub-tasks. The instance-level loss can be formulated as:
\begin{equation}
	\begin{aligned}
		L_{ins} = &-\lambda \sum_{i,j} [d_i logp_{i,j}^d + (1-d_i) log(1-p_{i,j}^d)] 
		\\&- (1-\lambda) \sum_{i,k} [d_i logp_{i,k}^r + (1-d_i) log(1-p_{i,k}^r)]
	\end{aligned}
	\label{lins}
\end{equation}

where $j\in \left \{1,...,K_1 \right \}$ and $k\in \left \{1,...,K_2 \right \}$. The source domain comprises $N_s$ images, while the target domain includes $N_t$ images. The balancing factor $\lambda$ is set as:
\begin{equation}
	\lambda = \sigma \left ( 4\cdot sign(N_t-N_s)\left(\frac{\max(N_s,N_t)}{\min(N_s,N_t)}-1 \right ) \right ) 
\end{equation}
where $\sigma$ is Sigmoid function employed for domain normalization. What's more, we introduce an L2-norm regularizer to maintain consistency between image-level and instance-level classifiers.

\subsection{Unlabeled Target Domain Training}
FOUS employs a prototype-relevance labeling method as a replacement for complex clustering algorithms, leading to a significant reduction in computational costs. In FOUS, the prototype vector needs to be initially selected or calculated, followed by the computation of the correlation between the prototype vector and all the samples. Then assign the label of the nearest prototype vector to the sample as a pseudo label. Therefore, the selection of the prototype vector is a critical aspect. And FOUS design two types of prototypes, namely the source-prototype vector and the random-prototype vector. Initially, the source-prototype vector is computed using well-labeled data from the source domain. Subsequently, to enhance FOUS's adaptability in the target domain, features are randomly selected from the target domain to create a random-prototype vector.

\textbf{Prototype vector initialization} We utilize the network $E_s(\cdot)$, pre-trained on the source domain, to extract features from the source domain samples. Subsequently, based on the ground-truth labels, the source-prototype vector can be computed as:
\begin{equation}
	P_k^s = \frac{1}{N_k} \sum_{f_i\in F_k^s} f_i
\end{equation}
where $F_k^s$ represents a set of features with the same labels $k$. $f_i$ is the feature of each instance. $N_k$ is the number of samples with the label $k$. The source-prototype vector $P^s$ is composed of $\left \{P_1^s,...,P_K^s \right \}$, $k\in [1,K]$. Particularly, $K$ means the number of identity categories in the source domain. However, models trained solely on pseudo-labels generated from source-prototype vectors tend to be sensitive to variations in the target domain and are susceptible to introducing a considerable amount of noise. Hence, FOUS sets up random prototype vectors in the target domain to enhance diversity. Similar to the source-prototype, $E_s(\cdot)$ is employed to extract features from target domain samples. Subsequently, $N_t$ vectors are randomly selected to form the random-prototype $P^t = \left \{P_1^t,...,P_{N_t}^t \right \}$.

\textbf{Labeling}  Once the source and random prototype vector are calculated, the labels for other sample instances in the target domain are determined by the k-Nearest Neighbor algorithm,
\begin{equation}
	l_i^s = \mathop{\arg\min}\limits_{l^s} dist(f_i,P^s)
	\label{lis}
\end{equation}
where $dist(a,b)$ means the Euclidean distance between $a$ and $b$. $f_i$ is the characteristic of the $i$-th instance. Eq.\ref{lis} signifies assigning the instance with the label of the vector in $P^s$ that is closest to $f_i$. All the labeled images collectively constitute a new dataset $D^s$. Similar to $D^s$, according to the random-prototype $P^t$, we can also derive a group of labels,
\begin{equation}
	l_i^t = \mathop{\arg\min}\limits_{l^t} dist(f_i,P^t)
	\label{lit}
\end{equation}

The assignment of pseudo labels through prototypes results in the creation of two datasets, which are $D^s = \left \{(x_i, l_1^s,...,l_n^s) \right \}_{i=1}^N$ and $D^t = \left \{(x_i, l_1^t,...,l_n^t) \right \}_{i=1}^N$, $n$ represents the number of instances in $x_i$. Then, the source-prototype is updated by the target datasets with pseudo labels,
\begin{equation}
	P_j^s = \frac{1}{N_j} \sum_{f\in F_j^t} f
\end{equation}
where $F_j^t$ means the features belonging to the cluster $j$. Therefore, FOUS does not use images from the source domain except for the first iteration.

\textbf{Label-flexible Training} To improve the training of models with flexible labels, FOUS constructs distinct memory structures to store features based on different label groups. Based on the source prototype $P^s$ and random prototype $P^t$, the pseudo labels $L^s$ and $L^t$ can be calculated. To mitigate the influence of noise, we designed two cluster-level memory structures to store the features at the training stage. Taking the $L^s$ as an example, the features $\left \{c_1,...c_K \right \}$ of each cluster are stored in a memory-based feature dictionary. $K$ means the number of clusters. After clustering with prototypes, the features in the cluster are utilized to initialize the memory dictionary, which could be expressed as,
\begin{equation}
	m_k = \frac{1}{N_k} \sum_{f\in F_k} f
\end{equation}
where $F_k$ represents the feature set of instances labeled as $k$, comprising all feature vectors in cluster $k$. $N_k$ is the number of features in $F_k$. When employing random-guided memory, we are provided with a query instance $q$ and perform feature comparisons with all features within the respective memory$\left \{m_1^t,...m_K^t \right \}$ structure utilizing the InfoNCE loss,
\begin{equation}
	L_c^t= -\log \frac{\exp(sim(f\cdot m_+^t)/\tau)}{ {\textstyle \sum_{k=1}^{K}}\exp(sim(f\cdot m_k^t)/\tau) }
	\label{lct}
\end{equation}
where $m_+^t$ denotes the features that share the same identity as f within the random-guided memory. $\tau$ is the proportional coefficient balancing factor. Similar to Eq.\ref{lct}, the loss $L_c^s$ of the source-guided memory also could be calculated.

In addition, FOUS employs invariant learning between images to bolster the model's generalization capability. An instance-level memory bank $\left \{m_1^l,...m_N^l \right \}$ is consequently established to store all features, and soft labels are estimated by evaluating the correlation between images.

Due to the absence of genuine labels, neighborhoods displaying high correlation play a role in diminishing the distance between similar images. Consequently, FOUS introduces an adaptive selection strategy. Taking inspiration from \cite{ding2020adaptive,ijcai2020p127,wang2022progressive}, a similarity threshold is defined. Only when the distance between the adaptively chosen neighbors and the provided image falls below this threshold, these samples are presumed to share the same label as the target image. Therefore, these samples should exhibit closer proximity to the target image. The corresponding loss can be formulated as:
\begin{equation}
	L_t^e = -\frac{1}{N} \sum_{j=1}^{N} \sum_{i=1,j \neq i}^{N} v_j^ilogp(i|m_j^t)
    \label{lte}
\end{equation}
where $v_j \in \left \{0,1 \right \}^N$. If $v_j^i = 1$, it means the $i$-th image is a reliable neighborhood of the $j$-th image. If $v_j^i = 0$, the $i$-th image is abandoned.

\textbf{Final loss} The final loss is composed of detection loss and re-id loss, which is formulated as,
\begin{equation}
	L = L_{ins}+L_c^t + L_c^s + L_t^e + L_s^e
\end{equation}
where $L_{ins}$ is calculated by Eq.\ref{lins}. $L_c^t$ and $L_c^s$ are calculated by Eq.\ref{lct}. $L_t^e$ and $L_s^e$ are calculated by Eq.\ref{lte}.

\section{Experiments}
\label{sec:exp}
\subsection{Datasets and Evaluation metrics}
\textbf{Datasets} The proposed method is evaluated on two large-scale benchmark datasets, CUHK-SYSU\cite{xiao2017joint} and PRW\cite{zheng2017person}. CUHK-SYSU consists of 12,490 images captured by real surveillance cameras and 5,694 images from movies and TV shows. The training set comprises 11,206 images and 5,532 query persons, while the test set includes 6,978 images and 2,900 query persons. The widely utilized PRW dataset comprises a total of 11,816 video frames captured by 6 cameras and 2,057 queries with 932 identities.

\textbf{Evaluation metrics} In the dataset configuration, annotations for the source domain dataset are available, while the ground truth boxes and identity information for the target domain dataset are not accessible. We employ widely used metrics such as mean average precision(mAP) and cumulative matching characteristic(CMC) top-1 accuracy for evaluating the re-identification subtask. For the detection task, average precision(AP) and recall rate are adopted as the evaluation metrics.

\subsection{Implementation Details}
Our experiments are implemented on NVIDIA GeForce GTX 3090Ti GPU with Pytorch\cite{paszke2017automatic}. FOUS utilize ResNet50\cite{he2016deep} pre-trained on ImageNet-1k\cite{deng2009imagenet} as the default backbone network. During the training process, the dimensions of input images are adjusted to 1500 $\times$ 900, and random horizontal flipping is applied for data augmentation. We employ the Stochastic Gradient Descent(SGD) method to optimize the model, with a mini-batch size set to 4 and a learning rate of 0.003. The momentum and weight decay are set to 0.9 and $5\times 10^{-4}$, respectively. For the k-nearest neighbor algorithm, the number of neighbors is set to 1. When PRW is chosen as the target domain, we pre-train on the source domain CUHK-SYSU for 8 epochs and subsequently engage in joint training for 10 epochs. Similarly, when CUHK-SYSU is chosen as the target domain, we pre-train on the source domain PRW for 2 epochs and subsequently engage in joint training for 14 epochs. However, the similarity between source domain samples is computed only once.

\begin{table}[]
	\centering
	\caption{Comparison of mAP($\%$) and rank-1 accuracy($\%$) with the state-of-the-art on CUHK-SYSU and PRW} 
	\label{Table1}
	\resizebox{0.5\textwidth}{!}{
		\begin{tabular}{cl|cc|cc}
			\hline
			\multicolumn{2}{c}{\multirow{2}{*}{Method}}                    & \multicolumn{2}{c}{CUHK-SYSU} & \multicolumn{2}{c}{PRW} \\ \cline{3-6} 
			\multicolumn{2}{c}{}                                           & mAP           & top-1          & mAP        & top-1       \\ \hline
			\multicolumn{1}{c}{\multirow{6}{*}{\rotatebox{90}{two-stage}}}   & DPM \cite{girshick2015deformable}         & -             & -              & 20.5       & 48.3        \\
			\multicolumn{1}{c}{}                             & MGTS \cite{chen2018person}        & 83.0          & 83.7           & 32.6       & 72.1        \\
			\multicolumn{1}{c}{}                             & CLSA \cite{lan2018person}        & 87.2          & 88.5           & 38.7       & 65.0        \\
			\multicolumn{1}{c}{}                             & IGPN \cite{dong2020instance}       & 90.3          & 91.4           & 47.2       & 87.0        \\
			\multicolumn{1}{c}{}                             & RDLR \cite{han2019re}       & 93.0          & 94.2           & 42.9       & 70.2        \\
			\multicolumn{1}{c}{}                             & TCTS \cite{wang2020tcts}       & 93.9          & 95.1           & 46.8       & 87.5        \\ \hline
			\multicolumn{1}{c}{\multirow{16}{*}{\rotatebox{90}{end-to-end}}} & OIM \cite{xiao2017joint}        & 75.5          & 78.7           & 21.3       & 49.4        \\
			\multicolumn{1}{c}{}                             & IAN \cite{xiao2019ian}         & 76.3          & 80.1           & 23.0       & 61.9        \\
			\multicolumn{1}{c}{}                             & NPSM \cite{liu2017neural}        & 77.9          & 81.2           & 24.2       & 53.1        \\
			\multicolumn{1}{c}{}                             & RCAA \cite{chang2018rcaa}       & 79.3          & 81.3           & -          & -           \\
			\multicolumn{1}{c}{}                             & CTXG \cite{yan2019learning}       & 84.1          & 86.5           & 33.4       & 73.6        \\
			\multicolumn{1}{c}{}                             & QEEPS \cite{munjal2019query}      & 88.9          & 89.1           & 37.1       & 76.7        \\
			\multicolumn{1}{c}{}                             & APNet \cite{zhong2020robust}      & 88.9          & 89.3           & 41.9       & 81.4        \\
			\multicolumn{1}{c}{}                             & HOIM \cite{chen2020hierarchical}       & 89.7          & 90.8           & 39.8       & 80.4        \\
			\multicolumn{1}{c}{}                             & NAE \cite{chen2020norm}        & 91.5          & 92.4           & 43.3       & 80.9        \\
			\multicolumn{1}{c}{}                             & NAE+ \cite{chen2020norm}       & 92.1          & 92.9           & 44.0       & 81.1        \\
			\multicolumn{1}{c}{}                             & AlignPS \cite{yan2021anchor}    & 94.0          & 94.5           & 46.1       & 82.1        \\
			\multicolumn{1}{c}{}                             & SeqNet \cite{li2021sequential}     & 93.8          & 94.6           & 46.7       & 83.4        \\ \hline
			\multicolumn{1}{c}{Ours}                         & FOUS        & \textbf{78.7}          & \textbf{80.5}           & \textbf{35.4}       & \textbf{80.8}        \\ \hline
		\end{tabular}
	}
\end{table}

\subsection{Comparison with State-of-the-Art Methods}
As there is currently limited interest in unsupervised pedestrian search methods, we begin by comparing FOUS with fully supervised methods in Table.\ref{Table1}, encompassing both two-stage and end-to-end methods. For the first time, we eliminate the need for computationally expensive clustering algorithms in assigning labels for weakly supervised and unsupervised pedestrian search tasks. Surprisingly, FOUS outperforms some supervised methods. The experimental results indicate a substantial gap between FOUS and state-of-the-art supervised methods. We aspire that our approach may serve as a catalyst for subsequent explorations in this field.
\begin{table}[]
	\centering
	\caption{Comparison of mAP($\%$) and rank-1 accuracy($\%$) with the weakly supervised and unsupervised methods on CUHK-SYSU and PRW} 
	\label{Table2}
	\resizebox{0.5\textwidth}{!}{
		
		\begin{tabular}{ccccc}
			\hline
			\multicolumn{1}{c|}{\multirow{2}{*}{Methods}}   & \multicolumn{2}{c|}{CUHK-SYSU}                                 & \multicolumn{2}{c}{PRW}                                       \\
			\multicolumn{1}{c|}{}                           & map                           & \multicolumn{1}{c|}{top-1}                         & map                           & top-1                         \\ \hline
			\multicolumn{1}{l}{\textit{weakly supervised:}} & \multicolumn{1}{l}{}          & \multicolumn{1}{l}{}          & \multicolumn{1}{l}{}          & \multicolumn{1}{l}{}          \\
			CGPS\cite{yan2022exploring}                                            & 80.0                          & 82.3                          & 16.2                          & 68.0                          \\
			R-SiamNet\cite{han2021weakly}                                       & 86.0                          & 87.1                          & 21.2                          & 73.4                          \\
			DICL\cite{jia2023collaborative}                                            & 87.4                          & 88.8                          & 35.5                          & 80.9                          \\
			SSL\cite{wang2023self}                                             & 87.4                          & 88.5                          & 30.7                          & 80.6                          \\ \hline
			\multicolumn{1}{l}{\textit{unsupervised:}}      & \multicolumn{1}{l}{\textit{}} & \multicolumn{1}{l}{\textit{}} & \multicolumn{1}{l}{\textit{}} & \multicolumn{1}{l}{\textit{}} \\
			DAPS\cite{li2022domain}                                            & 77.6                          & 79.6                          & 34.7                          & 80.6                          \\ \hline
			Ours                                            & \textbf{78.7}                          & \textbf{80.4}                          & \textbf{35.4}                 & \textbf{80.8}                 \\ \hline
		\end{tabular}
	}
\end{table}

We present a comparison of FOUS with state-of-the-art weakly supervised and unsupervised methods in Table.\ref{Table2}. Surprisingly, FOUS exhibits superior performance on the PRW dataset compared to all weakly supervised and unsupervised methods. For the CUHK-SYSU dataset, FOUS surpasses the existing unsupervised method DAPS by 1\% in mAP and 1.2\% in top-1, although it has not yet reached the level of performance achieved by weakly supervised methods. The reason may be as mentioned in DAPS. Since the number of images and identities in PRW is significantly less than in CUHK-SYSU, the detection performance of CUHK-SYSU as the target domain is poor, further leading to lower accuracy in the search task. The proposed attention-based domain alignment module enhances the quality of proposals, further bolstering the performance of detection in FOUS. Consequently, FOUS achieves a balance between reducing computational complexity, accelerating processing speed, and maintaining stable accuracy.

\begin{figure}[tbp!]
	\centering
	\setlength{\belowcaptionskip}{-1mm}
	\vspace{-0.10cm} 
	\subfigtopskip=-1pt 
	\subfigbottomskip=-1pt 
	\subfigcapskip=-5pt 
	\subfigure[$N_t$]{\label{f2a}
		\includegraphics[width=1.55in]{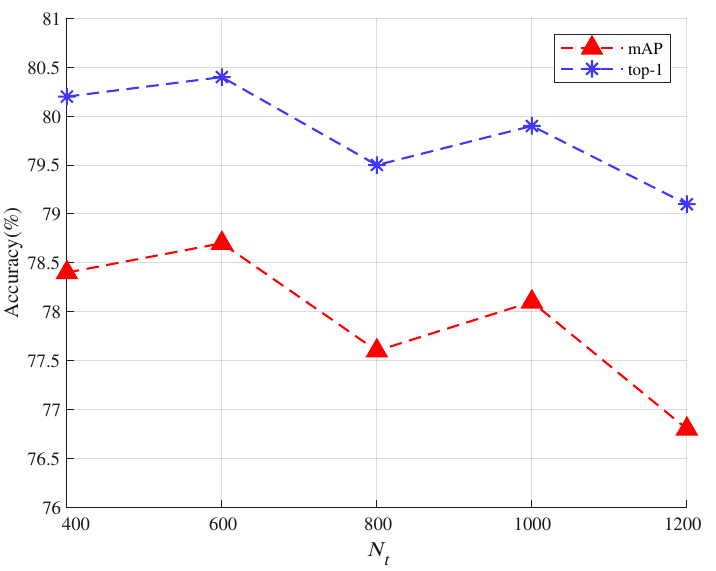}}
	\quad    
	\subfigure[eopch]{\label{f2b}
		\includegraphics[width=1.55in]{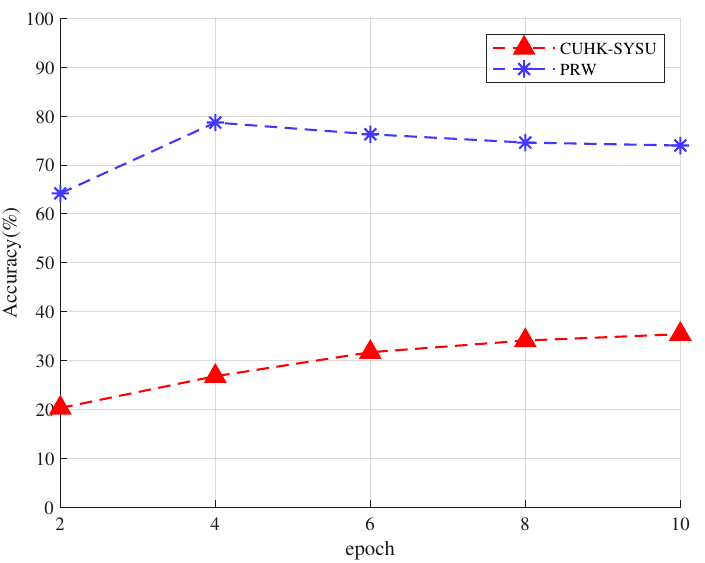}}	
	
	\caption{\subref{f2a} represents the influence of varying the number of random prototypes on the map and top-1 metrics on the CUHK-SYSU dataset. \subref{f2b} illustrates the impact of different pre-training rounds on two datasets. }
	\label{GS}
\end{figure}

\begin{table}[]
	\centering
	\caption{Validating the effectiveness of different modules on CUHK and PRW datasets. AM: Attention Mechanism. TDA: Task-sensitive Domain Alignment. LFT: Label-flexible Training} 
	\label{Table3}
	\begin{tabular}{ccc|cc|cc}
		\hline
		\multirow{2}{*}{AM} & \multirow{2}{*}{TDA} & \multirow{2}{*}{LFT} & \multicolumn{2}{c|}{CUHK-SYSU} & \multicolumn{2}{c}{PRW} \\
		&                      &                      & map               & top-1             & map           & top-1          \\ \hline
		&         &                      & 50.4              & 52.8              & 26.7              & 74.6               \\
		$\checkmark$        & $\checkmark$         &                      & 68.1              & 71.6              & 33.2              & 80.4               \\
		& $\checkmark$         & $\checkmark$         & 64.3              & 66.9              & 28.0              & 77.1               \\
		$\checkmark$        &                      & $\checkmark$         & 72.3              & 75.7              & 32.9              &  80.2              \\
		$\checkmark$        & $\checkmark$         & $\checkmark$         & \textbf{78.7}              & \textbf{80.4}              & \textbf{35.4}              & \textbf{80.8}               \\ \hline
	\end{tabular}
\end{table}
\subsection{Ablation Study}
In this subsection, we perform ablation experiments to validate the significance of each component in FOUS. In Table.\ref{Table3}, we compare the combination of baseline methods with different proposed modules and report the results on both CUHK-SYSU and PRW datasets. The baseline model achieves 50.4\% mAP and 52.8\% Top-1 accuracy on the CUHK-SYSU dataset. FOUS implemented three components and achieved a 28.2\% mAP and 28.0\% top-1 improvement over the baseline. We find that if only use the prototype-guided labeling method to generate coarse label groups without utilizing the label-flexible network to refine the labels, the accuracy will decrease significantly by 10\% on CUHK-SYSU and 7\% on PRW. If the attention mechanism is abandoned to enhance proposals, the accuracy drops even further to 14.3\% on CUHK-SYSU. Furthermore, neglecting the domain difference between the two subtasks leads to a reduction in accuracy by 6.3\% on CUHK. Hence, the three components we proposed effectively enhance the performance and generalization ability of the model.

Furthermore, to verify the effectiveness of the designed attention module, we compared it with state-of-the-art attention methods. As shown in Table.\ref{Table5}, our proposed attention module outperforms significantly, achieving an average accuracy 1.8 higher than CBAM, ECAM, SGEM, and GCT methods on CUHK-SYSU. We can confidently infer that global information aggregation and interaction with proposals are beneficial for model learning.

Last but not least, we compare with state-of-the-art weakly supervised and unsupervised methods in terms of network parameter amount and computational effort. Table.\ref{Table4} presents the comparative results concerning computational requirements, parameter count, training time per iteration, and accuracy during the process. As FOUS computes the similarity of source domain samples only once, it reduces computational costs by nearly half compared to DAPS, which utilizes clustering algorithms. This reduction is achieved at a minor cost in terms of parameters. Furthermore, training time is relatively reduced by 25\%, and there is a remarkable improvement in accuracy rather than a decline. We have abandoned the complex clustering method and, instead, proposed a prototype correlation labeling algorithm, resulting in a significant reduction in the number of parameters and computational workload.

\subsection{Parameter analysis}
Setting several parameters in FOUS may affect the performance of the model. Therefore, we evaluate these parameters in our paper, such as the number of random-prototype vectors and the pre-trained epochs during the training stage. The results on CUHK-SYSU and PRW under different settings are shown in Fig.\ref{GS}.

\textbf{Analysis of the random selected prototype number}
As shown in Fig.\ref{f2a}, we vary the number of random prototype vectors $N_t$ to 400, 600, 800, 1000, and 1200 on CUHK-SYSU, respectively. And $N_t$ is set to 100, 200, 300, 400, 500 on PRW, respectively. In detail, maintaining the parameters of batch size and pretraining rounds unchanged, as illustrated in Figure 3, the model performance varies with the different settings of $N_t$. This underscores the significance of randomly selecting the number of prototypes for generating pseudo-labels. What's more, Fig.\ref{f2a} also shows that $N_t$ = 600 and $N_t$ = 300 can achieve the best accuracy on CUHK-SYSU and PRW, respectively.

\textbf{Analysis of the pre-trained epochs}
In this part, the paper explores the relations between the pre-trained epochs and the whole model performance. As observed in Fig.\ref{f2b}, when CUHK-SYSU is embraced as the target domain, the performance peaks at a pre-trained epoch of 10, while with epoch = 4 for PRW. Our analysis of this result is that in smaller source domains, even limited additional target information may contribute to cross-domain generalization. In contrast, for larger source datasets, unreliable target suggestions may have a negative impact on mitigating domain differences.

\begin{table}[]
	\centering
	\caption{Comparing the network parameters and computational requirements with weakly supervised and unsupervised methods on PRW.} 
	\label{Table4}
	\resizebox{0.5\textwidth}{!}{
		
		\begin{tabular}{c|ccccc}
                \hline
			Methods   & GFLOPS & PRARMS & Time & map  & top-1 \\ \hline
			CGPS      & 281    & 49.2M  & 1.1h   & 16.2 & 68.0  \\
			R-SiamNet & 320    & 52.5M  & 1h   & 21.4 & 75.2  \\
			DAPS      & 628    & 53.4M  & 1.6h  & 34.7 & 80.6  \\ \hline
			Ours      & \textbf{388}    & \textbf{53.6M}  & \textbf{1.2h}   & \textbf{35.4} & \textbf{80.8}  \\ \hline
		\end{tabular}
	}
\end{table}

\begin{table}[]
	\centering
	\caption{Validating the effectiveness of different attention mechanism on CUHK and PRW datasets.} 
	\label{Table5}
	\resizebox{0.5\textwidth}{!}{
		\begin{tabular}{c|cc|cc}
			\hline
			\multirow{2}{*}{\begin{tabular}[c]{@{}c@{}}Attention \\ Models\end{tabular}} & \multicolumn{2}{c|}{CUHK-SYSU} & \multicolumn{2}{c}{PRW} \\& map               & top-1             & map           & top-1          \\ 
			\hline
			CBAM\cite{woo2018cbam}                                                                            & 77.2              & 79.0              &  33.6             & 79.4               \\
			ECAM\cite{wang2020eca}                                                                            & 76.2              & 78.1              &  30.1             & 78.6               \\
			SGEM\cite{li2019spatial}                                                                            & 77.0              & 78.7              &  32.3             & 80.1               \\
			GCT\cite{ruan2021gaussian}                                                                             & 76.8              & 77.9              &  32.6             & 80.2               \\ \hline
			Ours                                                                            & \textbf{78.7}              & \textbf{80.4}              & \textbf{35.4}              &  \textbf{80.8}              \\ \hline
		\end{tabular}
	}
\end{table}

\section{Conclusion}
\label{sec:con}
In this paper, we present the FOUS framework designed for the domain adaptive pedestrian search task. FOUS introduces two components, one is a prototype-guided labeling algorithm, which replaces the commonly used complex clustering algorithms in this domain, significantly reducing computational costs at the expense of accuracy. The other is an attention-based domain alignment module, alleviating the impact of low-quality candidate boxes on re-identification tasks in unsupervised detection. Together, these components complement each other, reducing computational overhead while enhancing the model's performance. Judging from the results, experiments on two foundational datasets demonstrate the performance of FOUS and the effectiveness of the introduced modules and achieve state-of-the-art performance.

\bibliographystyle{named}
\bibliography{ijcai24}

\end{document}